\documentclass[a4paper, 10pt, conference]{ieeeconf}      

\IEEEoverridecommandlockouts                              

\usepackage{graphics} 
\usepackage{epsfig} 
\usepackage{mathptmx} 
\usepackage{times} 
\usepackage{amsmath} 
\usepackage{amssymb}  
\usepackage{booktabs}
\usepackage{multirow}
\usepackage{bbding}
\usepackage{makecell}
\usepackage{color}
\usepackage{amsfonts}
\usepackage{mathtools}
\usepackage{subcaption}

\DeclareMathAlphabet{\mathcal}{OMS}{cmsy}{m}{n}
\usepackage{algorithm}
\usepackage{algpseudocode}
\usepackage{hyperref}

\usepackage{authblk}

\title{\LARGE \bf
Analyzing and Enhancing Closed-loop Stability in Reactive Simulation
}

\author{{Wei-Jer Chang, Yeping Hu, Chenran Li, Wei Zhan, and Masayoshi Tomizuka}\thanks{
*This work was supported by Hong Kong Centre Logistics Robotics. \\
 W.J. Chang, Y. Hu, C. Li, W. Zhan and M. Tomizuka are with the Department of Mechanical
Engineering, University of California, Berkeley, CA 94720 USA
[weijer chang, yeping hu, 	
chenran li, wzhan, tomizuka@berkeley.edu]}} 

\begin{document}

\maketitle 
\thispagestyle{empty}
\pagestyle{empty}

\begin{abstract}
Simulation has played an important role in efficiently evaluating self-driving vehicles in terms of scalability. Existing methods mostly rely on heuristic-based simulation, where traffic participants follow certain human-encoded rules that fail to generate complex human behaviors. Therefore, the reactive simulation concept is proposed to bridge the human behavior gap between simulation and real-world traffic scenarios by leveraging real-world data. However, these reactive models can easily generate unreasonable behaviors after a few steps of simulation, where we regard the model as losing its stability. To the best of our knowledge, no work has explicitly discussed and analyzed the stability of the reactive simulation framework. In this paper, we aim to provide a thorough stability analysis of the reactive simulation and propose a solution to enhance the stability. Specifically, we first propose a new reactive simulation framework, where we discover that the smoothness and consistency of the simulated state sequences are crucial factors to stability. We then incorporate the kinematic vehicle model into the framework to improve the closed-loop stability of the reactive simulation. Furthermore, along with commonly-used metrics, several novel metrics are proposed in this paper to better analyze the simulation performance.

\end{abstract}

\section{INTRODUCTION}
  
Self-driving vehicles have the potential to build efficient and safe transportation in society. To enable self-driving vehicles to safely and efficiently navigate in human-involved environments, the design of interactive algorithms for different modules (e.g., prediction, decision-making) is of great importance. There are three general approaches to evaluate the overall performance of self-driving vehicles: road-testing, heuristic-based simulation, and log replay. Road testing is the most direct way to evaluate the performance of autonomous vehicles in the real world. Nevertheless, the testing process is expensive and time-consuming. In addition, it is hard to directly evaluate the performance of autonomous vehicles under safety-critical scenarios in the real world. For heuristic-based simulation, traffic participants will follow specific human-encoded or traffic rules. Although heuristic-based simulation can mimic basic driving behaviors, it is still challenging to generate irregular and complex human behaviors. For log replay approach, traffic participants will follow their recorded actions to imitate real-world scenarios. However, traffic participants that are operated by the log replay cannot react to the self-driving vehicle's new actions that differ from its recorded actions. For example, when a self-driving vehicle determines to slow down (new action) instead of maintaining the same velocity (recorded action) in the simulation, it might collide with any simulated vehicles behind.  
To deal with the aforementioned problems, several recent works proposed the concept of reactive simulation \cite{TrafficSim}\cite{simnet} to tackle the challenge of bridging the human behavior gap between simulation and real-world scenarios. The term 'reactive simulation' means traffic participants in the simulation environment will react to the self-driving vehicle similar to what human drivers will do in the real world. Each traffic participant, or so-called simulated agent, is simulated by reactive simulation, and the self-driving vehicle is the vehicle we want to evaluate in the simulation environment. To simulate realistic human behaviors, reactive simulation is usually implemented by using a predictor that acts as a controller to control each simulated agent's actions. However, as we will show later in our experiments, if a predictor is directly used in a closed-loop setting to consequently generate simulated actions, simulated traffic participants will generate undesirable behaviors. For example, the simulated traffic participants will easily drive off-road or drive with jerky trajectories after few steps of iteration. We regard such model performance as a lack of stability. To the best of our knowledge, no work has explicitly discussed and analyzed the stability of the reactive simulation frameworks. In this paper, an in-depth stability analysis of the reactive simulation and a possible solution to improve stability are provided. Our contributions are summarized as follows:  
\begin{enumerate}
    \item Incorporate the kinematic vehicle model into the reactive simulation framework.
    \item Provide a thorough stability analysis of the reactive simulation framework and discuss the effects of the kinematic vehicle model.
    \item Propose several novel metrics to analyze the simulation performance better.
  
\end{enumerate}

The remainder of this paper is structured as follows. The related works are presented in section II. The closed-loop reactive simulation framework is explained in Section III. In Section IV, experimental results and ablation study are provided and discussed. Section V concludes the paper.

\section{Related works}

\subsection{Scenario Generation}
Scenario generation often uses generative models, such as Generative Adversarial Network (GAN) and Variational Autoencoders (VAE), to simulate driving scenarios in autonomous driving \cite{20}\cite{21_CMTS}. In addition, some controllable generation methods \cite{Diverse}\cite{SceGene} are proposed recently. \cite{Diverse} proposed a style generative model that can generate diverse interactions with different safety levels by altering style coefficients and SceGene \cite{SceGene} presented a bio-inspired traffic scenario generation method in a controllable manner. However, scenario generation only simulates the interactions based on historical states, and it cannot simulate the reactive behaviors during the simulation.
\subsection{Reactive Behavior Simulation}
There are two different reactive behavior simulation approaches: heuristic-based and learning-based simulation. In the heuristic-based simulation, the traffic participants are controlled by specific human encoded rules. A general lane-changing car-following behavior model is proposed in \cite{behavioural_car_follow}. The intelligent driver model (IDM) \cite{Congested} is a continuous microscopic car-following model that has smooth braking and acceleration strategies. An overview of microscopic traffic modeling and simulation techniques is provided in \cite{Agents_traffic_sim}. Recently, high-fidelity simulators such as CARLA \cite{CARLA} and SUMO \cite{SUMO} have been widely used to evaluate self-driving vehicle algorithms. Although the heuristic-based approach can produce reasonable trajectories, it loses the diversity to capture human behavior since human behavior is complex and hard to describe by hand-crafted rules. For example, it is difficult to describe complex behavior when human drivers negotiate during merging scenarios. 

As a result, several learning-based reactive simulation works were proposed to capture realistic multi-agent behaviors \cite{TrafficSim}\cite{simnet}. Both TrafficSim \cite{TrafficSim} and SimNet \cite{simnet} demonstrated the reactive simulation concept, which uses a neural network as a predictor to control the traffic participants at each time step, and each participant will "react" to each other every time step. The main advantage of the reactive simulation is that every agent's new action will be affected by other agents' motion states to model the interaction. Usually, reactive simulation is implemented by the "closed-loop" concept: the future states are continuously predicted based on the new states to simulate how agents will interact with each other in the next few seconds. In addition, TrafficSim demonstrated the usage of simulations as an effective data augmentation approach for training a better motion planner, and SimNet utilized the simulation model to evaluate an existing motion planner. 
The main difference between scenario generation and reactive simulation is that scenario generation will generate the driving scenarios directly based on historical states (open-loop). On the other hand, in reactive simulation, the simulated results will be based on the interaction between the self-driving vehicle and the traffic participants at each time step (closed-loop). 

\begin{figure*}
    \centering \includegraphics[width=\linewidth]{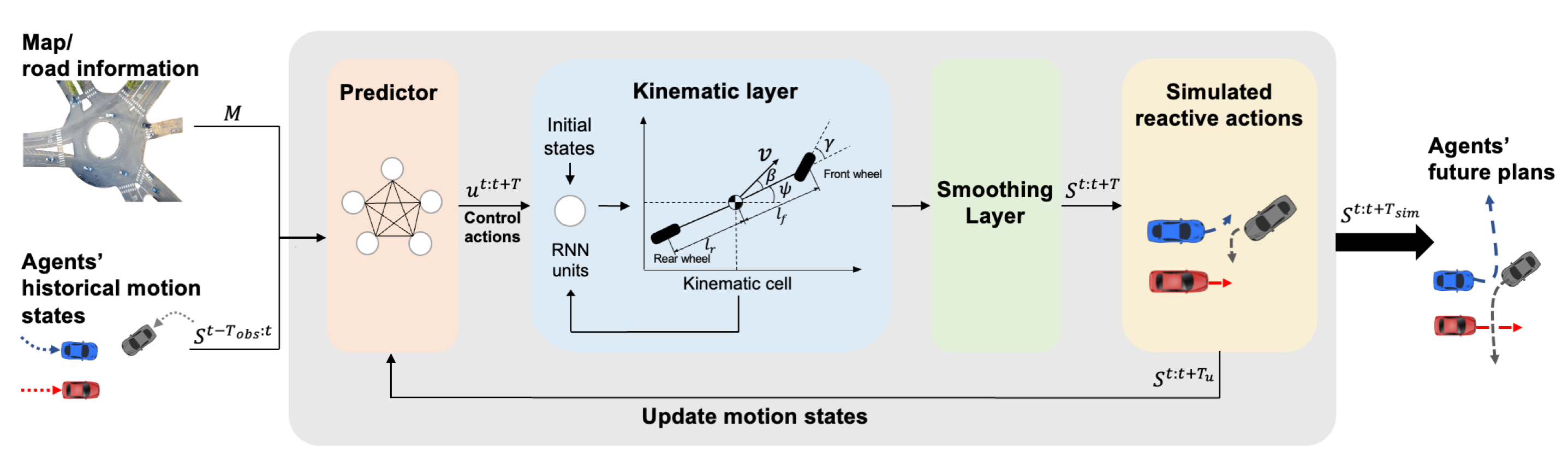}
    \caption{Visualization of our proposed closed-loop simulation framework. Based on map information and historical states, (a) the predictor backbone will first predict agents' control actions. Then, the control actions are fed into (b) the kinematic layer to generate kinematically feasible trajectories. At last,  the smoothing layer will produce relatively smooth trajectories. In each iteration, each agent will follow their simulated reactive actions, and their motion states will be updated. The framework will iteratively predict agents' future trajectories based on the new states to simulate agents' future plans.
    }
    \label{Simulation_framework}
\end{figure*}


\subsection{Kinematics in Motion Prediction}
Accurately predicting the motion and interaction of the surrounding vehicles are necessary capabilities of autonomous vehicles. In recent years, machine learning methods have greatly improved the capability of predicting the traffic participants in different driving scenarios \cite{SIMP}\cite{causal}\cite{SGN}. Nonetheless, pure machine learning methods cannot guarantee the dynamic feasibility \cite{kinematic_model} of trajectories. For example, the predicted vehicle may turn in a small radius that does not satisfy the non-holonomic constraints of the vehicle. Therefore, the concept of adding a kinematic layer to machine learning methods was proposed \cite{Ma2019WassersteinGL}\cite{Deep_kinematic_model}. \cite{Ma2019WassersteinGL} proposed a novel generative model structure to enable neural networks to satisfy the kinematic constraints automatically, and \cite{Deep_kinematic_model} combined a deep CNN model with the kinematic bicycle model for kinematic feasibility guarantees. Recently, several works have incorporated planning methods in neural networks to generate physically feasible trajectories \cite{Predict_Vehicle_Trajectories_model} \cite{Physically_feasible}. \cite{Predict_Vehicle_Trajectories_model} proposed a model-based generator to generate future trajectories under explicit constraints and used a learning-based selector to capture implicit interactions. Compared to adding a kinematic layer, these planning-based methods can also include environmental constraints \cite{Predict_Vehicle_Trajectories_model}, but they may lose the complexity of capturing short-term human behaviors.

\section{Closed-loop simulation framework}

In this section, we present our closed-loop simulation framework, as shown in Fig. 1. Closed-loop simulation means the simulator will iteratively simulate traffic participants' future plans based on the new simulated states. The framework can be regarded as a controller that controls simulated agents' motion. The main motivation behind the closed-loop concept is to simulate the interaction between the traffic participants in every time step. Our framework mainly consists of three components: a backbone predictor, a kinematic layer, and a smoothing layer. The predictor will first predict agents' control actions based on map information and historical trajectories. Then, the control actions are fed into the kinematic layer to generate feasible trajectories. At last, the smoothing layer will produce relatively smooth trajectories.  

\begin{figure}[htbp]
\centering
\minipage{0.5\textwidth}
  \includegraphics[width=1\linewidth]{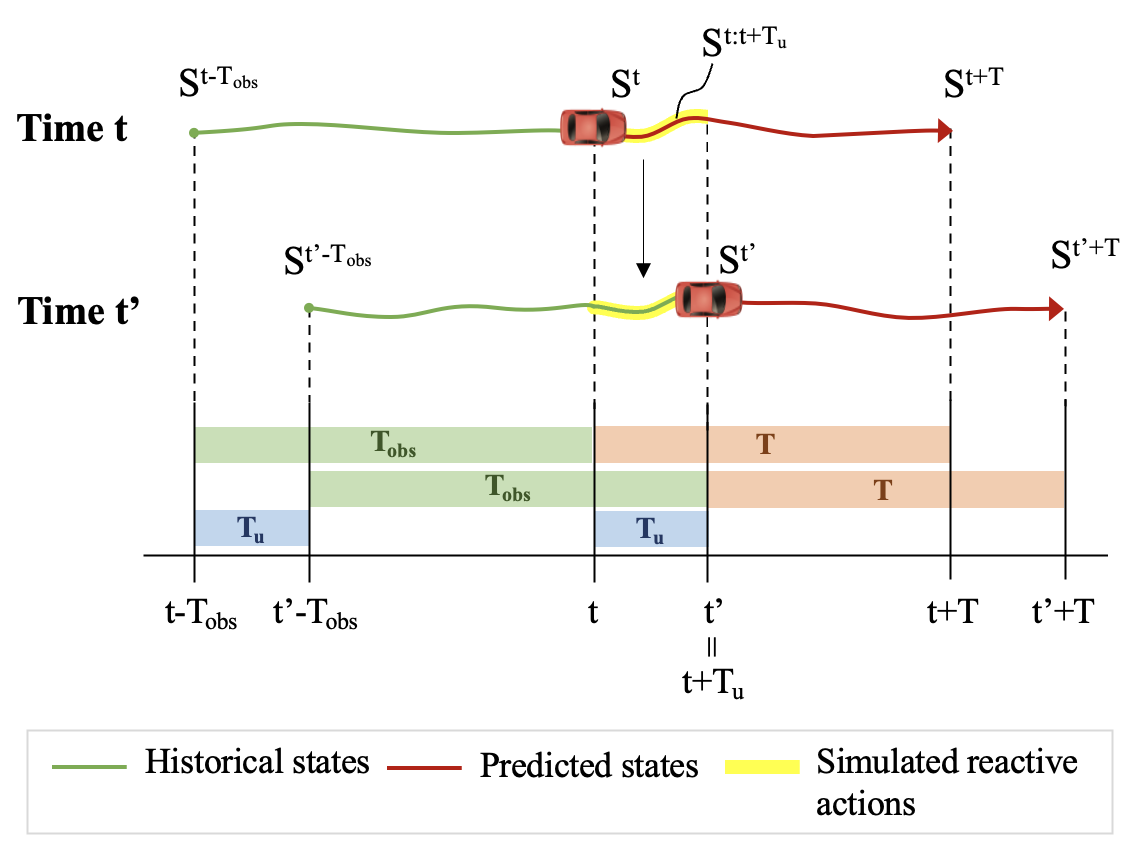}
  \caption{Illustration of the updating strategy of closed-loop simulation. At time $t$, the model will predict the predicted agent's states $S^{ t:t+T}$. Then, the closed-loop simulation framework will update the simulated reactive actions $S^{t:t+T_{u}}$.}
\endminipage\hfill
\end{figure}

\subsection{Closed-loop Formulation}
Based on map information $M$ and agents' historical motion states $S^{t-T_{obs}:t}$, we aim to simulate agents' future states in the next $T_{sim}$ time steps $S^{ t:t+T_{sim}}$. Here, $T_{obs}$ denotes the time length of observed historical states and $t$ is the current time step. Specifically, at a given time step $t$, $S^{t} = {\left\{ s_{1}^{t},s_{2}^t,...,s_{N}^t\right\}}$ represents all $N$ agents' dynamic states, and the $k$-th actor's state $s_{k}^{t}$ contains its position $\left(x_{k}^{t}, y_{k}^{t}\right)$, heading angle $\psi_{k}^{t}$, velocity $v_{k}^{t}$, vehicle length $l_{k}$, and vehicle width $w_{k}$. The updating strategy in closed-loop simulation is shown in Fig. 2. At each iteration, the model will predict simulated agents' future states $S^{t:t+T}$ in the next $T$ time steps. Then, the simulated agents will follow their simulated reactive actions $S^{t:t+T_{u}}$ depending on the updated time length $T_{u}$. Therefore, the new historical states $S^{t'-T_{obs}:t'}$ at time $t'$ in the next iteration will be the concatenation of past historical states and the new states $S^{t'-T_{obs}:t'} = \left[S^{t+T_{u}-T_{obs}:t}, S^{t:t+T_{u}} \right]$. 
 The simulation framework can be served as an evaluation platform for motion planners. In the simulation environment, the ego vehicle is controlled by a motion planning algorithm, while other agents are closed-loop controlled by the simulator \cite{simnet}. The main advantage of this formulation is that we can construct highly interactive and challenging scenarios to improve testing efficiency. In addition, it can also enhance machine learning models' prediction performance from a data augmentation perspective.
 
\subsection{Predictor}
In this paper, we utilize Vectornet \cite{Vectornet} as our backbone predictor in the overall simulation framework, which is treated as a controller to simulate each agent's future behaviors. Instead of using raster-based images to represent map information, Vectornet proposed vectorized representation to encode both static HD map information and agents' dynamic states as sequences of vectors and used Graph Neural Network (GNN) to model interactions between different entities. Note that the backbone predictor module can be replaced by any other prediction model.

\subsection{Kinematic Layer}
During the closed-loop simulation, we observed that if the predicted trajectories had a small discrepancy between past trajectories, the model would easily generate unreasonable behaviors, such as driving off-road or driving with jerky trajectories. The detailed discussions can be found in Section IV. Therefore, we use a kinematic layer \cite{Deep_kinematic_model} to generate kinematically feasible trajectories by following the kinematic bicycle model relation from the paper \cite{kinematic_model}:

\begin{equation}
\begin{aligned}
    \dot x &= v\cos \left(\psi +\beta\right), \\
    \dot y &= v\sin \left(\psi +\beta\right),  \\
    \dot \psi &= \frac{v}{l_{r}} \sin\left(\beta\right),  \\
    \dot v &= a, \\
    \beta &= {tan}^{-1}{\left( \frac{l_{r}}{l_{r}+l_{f}}\tan{\gamma} \right)}
\end{aligned}
\end{equation}

The relation of kinematic bicycle model parameters is shown in Fig. 1. $\gamma$ is the steering angle, and $l_{r}$ and $l_{f}$ is the distance from the center to the rear wheel and front wheel, respectively. $\beta$ is the angle of current velocity $v$ of the center of mass with respect to the vehicle's longitudinal axis. $a$ denotes the acceleration of the vehicle. The hyper-parameter $l_r$ in Eq. (1) denotes the distance from the rear wheel to the center of the vehicle. If not provided in the data, $l_r$ can be estimated by the historical states of the vehicle. In our settings, we use the predictor backbone to estimate the ratio of $\frac{l_r}{l_r+l_f}$ and calculate $l_r$ based on the vehicle's length ${l_r+l_f}$ from data. 

In order to output a sequence of actions, we utilize the aforementioned equations to define a Recurrent Neural Network (RNN) unit in the kinematic layer. Instead of directly outputting agents' trajectories, we let the last layer of the neural network output control parameters $u_{t} = [a_{t},\beta_{t}]$. Note that we use $\beta_{t}$ instead of $\gamma_{t}$ as the second input since $\beta_{t}$ and $\gamma_{t}$ have static one-to-one relationships, and we noticed an improvement of the model performance using $\beta_{t}$. By taking the last frame of historical states as the initial states and the control parameters at every frame $t$, the RNN unit will roll out each agent's future trajectories and update its states. This approach can guarantee that the model will output kinematically feasible trajectories. In addition, the kinematic layer also incorporates the time dependencies between each frame in the model.  One of the main advantages of the kinematic layer is that it can not only predict 2D positions $\left(x^{t}, y^{t}\right)$, but also predict velocity $v^{t}$ and heading angle $\psi^{t}$ at the same time. Such characteristic enables direct operations of the closed-loop simulation without estimating velocity $v^{t}$ and heading angle $\psi^{t}$ based on 2D positions. All of the parameters in the kinematic layer are non-trainable, and we can regard the layer as nonlinear functions that decode agents' trajectories. The backbone predictor combined with the kinematic layer can be viewed as predicting control parameters in the kinematic bicycle model space instead of the x-y space.

The reason we want to incorporate the kinematic bicycle model into the framework is to ensure the feasibility and smoothness of predicted trajectories, and we will discuss the importance of smooth trajectories in the closed-loop simulation. With the added kinematic layer, we can scale the output range of the predictor to bound the acceleration and angle in reasonable ranges. Besides using the kinematic bicycle model relation, we also applied kinematic motion equations of particles in the kinematic layer with $a_x$ and $a_y$ as the control parameters, which we called the simplified model kinematic layer. Using kinematic motion equations of particles can be regarded as simplifying the vehicle to a point mass. We then use the accelerations $a_x$ and $a_y$ to roll out the agent's trajectories with the same RNN concept. This simplified kinematic layer seems to have the best performance in terms of stability among three different settings: predictor without kinematic layer, predictor with kinematic bicycle layer, and predictor with simplified kinematic layer, and we will provide a detailed analysis in the experiment section.

\subsection{Smoothing Layer: Weighted Average}
To prevent sudden behavior switches during the closed-loop simulation process, we propose a weighted average smoothing layer to eliminate trajectory's discontinuity when predicted states are added to the input during closed-loop simulation. We denote the weighted average trajectory at time $t$ as $\left[x^{t:t+T},y^{t:t+T} \right]_{w}^{T}$ and the new predicted trajectory at time $t'$ as $\left[x^{t':t'+T},y^{t':t'+T} \right]_{p}^{T}$. We use Eq. (2) to update the weighted average trajectory $\left[x^{t:t+T},y^{t:t+T} \right]_{w}^{T}$. Then, the velocity and heading angle are derived based on the weighted average trajectory. This method can mitigate the discrepancy between the training and testing environment, which is effective in ensuring the stability of the closed-loop simulation.

\begin{equation} \label{eq:1}
    \begin{bmatrix} x^{t':t'+T} \\ y^{t':t'+T}\end{bmatrix}_{w} = (1-\alpha)\begin{bmatrix} x^{t':t'+T} \\ y^{t':t'+T}\end{bmatrix}_{p} + \alpha\begin{bmatrix} x^{t:t+T} \\ y^{t:t+T}\end{bmatrix}_{w}
\end{equation}

\section{Experiments}
In this section, we first showcase the failure cases by directly using a predictor to predict 2D positions in the closed-loop simulation framework. We showcase both visualization and quantitative analysis in the ablation study for the kinematic layer and the smoothing layer in closed-loop simulation. The sensitivity issue of the kinematic layer and the effect of the weighted average smoothing layer are discussed.

\subsection{Dataset}
We train and test our model using the USA\_Roundabout\_FT data from the INTERACTION dataset \cite{interactiondataset}, which contains highly interactive driving cases. As shown in Fig. 1, the roundabout has six branches, and each branch has two directions (in and out), which creates more interactions in the scenario. In this paper, we aim to test the closed-loop simulation approach and evaluating the framework's performance under this challenging scenario.

\subsection{Implementation Details}
For the predictor, we vectorized the HD map and past 1-second (10 frames) agent trajectories and trained the VectorNet backbone to predict the next 3 seconds of the target agent's trajectory. In the simulation, the simulated agent has access to other agents' states within an observable radius of 70 meters. For the simulation setup, we use 1-second ground truth data to initialize the agent's states and unroll the simulation for 5 seconds. We set 5-second as the simulation duration time since most of the interaction scenarios covered in the dataset happen in this range. The updated frequency is 10 Hz, which means we simulate 0.1 seconds in every iteration. At each iteration, the simulated agent will take the control action at the first step, and the new states will be the next input to the closed-loop simulation framework.

To thoroughly analyze the closed-loop simulation framework, we set up three different framework settings: predictor without kinematic layer, predictor with kinematic bicycle layer, and predictor with simplified kinematic layer. Each setting is composed of two sub-settings: with and without a weighted average method.
A total of six different framework settings are considered, which details are shown below:
\begin{itemize}
  \item \emph{xy} : Directly use Vectornet as the predictor to simulate future trajectories.
  \item \emph{xy weighted}: Directly use Vectornet with a weighted average smoothing layer to simulate future trajectories.
  \item \emph{kinematic} : Use Vectornet to predict control parameters $u_{t} = [a_{t},\beta_{t}]$, which will be fed into the kinematic layer with the kinematic bicycle model equations to simulate future trajectories.
  \item \emph{kinematic weighted}: Use Vectornet to predict control parameters $u_{t} = [a_{t},\beta_{t}]$, which will be fed into the kinematic layer with the kinematic bicycle model equations to simulate future trajectories with weighted average smoothing.
  \item \emph{axay} : Use Vectornet to predict control parameters $u_{t} = [a_{x},a_{y}]$, which will be fed into the kinematic layer with the particle acceleration equations to simulate future trajectories.
  \item \emph{axay weighted}: Use Vectornet to predict control parameters $u_{t} = [a_{x},a_{y}]$, which will be fed into the kinematic layer with the particle acceleration equations to simulate future trajectories with weighted average smoothing.
  
\end{itemize}

Note that to run the closed-loop simulation, we need to calculate the updated motion state from the framework output. For \emph{xy} and \emph{xy weighted}, we use Cubic spline \cite{cubic_spline} to interpolate the trajectories and use the interpolated function to derive heading angle and velocity. Similarly, for \emph{axay} and \emph{axay weighted}, we obtain the heading angle from Cubic spline interpolation, while the updated velocity is directly obtained from the kinematic layer. To analyze the model's performance, we simplify the simulation process by only simulating one car in each case and letting the other traffic participants follow their recorded actions. For the smoothing layer, we choose $\alpha = 0.2$. 

\subsection{Metrics}
Since there are no standard evaluation metrics for closed-loop simulation, we follow the metrics defined in \cite{TrafficSim}\cite{simnet} while proposing motion smoothness (MS) and predicted trajectory difference (TD) to analyze the stability of the model performance.


\textbf{Collision Rate (CR):} Collision rate reflects partial interaction of how the simulated agent reacts to other traffic participants. We define collision rate as the number of times the simulated agent collides out of the total number of sampled scenarios. The simulated agent is considered to have a collision if its bounding box overlaps with that of the other agents.

\textbf{Simulation Realism:} Average distance error (ADE) and final distance error (FDE) between ground truth and simulated positions are reported to measure how well the simulation can reconstruct real-world traffic scenarios. 

\textbf{Motion Smoothness (MS):} During the closed-loop simulation, the smoothness is strongly related to stability. Therefore, we use the absolute value of jerk, the rate that the agent's acceleration changes with respect to time to measure the smoothness of simulated trajectories.

\textbf{Trajectory Difference (TD):} Trajectory difference is defined as the mean square error between the predicted trajectories at a different time step. Instead of the above closed-loop metrics, we also define open-loop metrics (i.e., how the prediction changes at each iteration). Specifically, in our setting, we predict 30 frames at each time step, so the difference is the overlapped 29 frames between the two predictions. Trajectory difference measures the consistency of the model when unrolling in closed-loop simulation. Usually, an abrupt change in the ego vehicle's behavior is abnormal and will cause the predictor to generate unreasonable outputs.

\textbf{Prediction metrics:} We also provide the prediction metrics of each model, including average distance error (ADE), final distance error (FDE), and Miss Rate (MR). We consider a case as missed when the difference between the predicted vehicle's position and the ground truth position at the final frame is larger than the given lateral or longitudinal threshold defined in \cite{interactiondataset}. MR is the percentage of missed cases over all cases. 


\subsection{Analysis of Failure Cases}
In this subsection, we demonstrate how directly using the predictor would cause the model to fail easily during the closed-loop simulation. In fact, when the simulated ego vehicle continuously follows its predicted trajectories, the discontinuity of the predicted states will cause the model to yield unreasonable trajectories.

When the simulated agent is entering the roundabout shown in Fig. 3(a), we can observe that the ego vehicle starts to deviate at $t = 1.2s$. According to the velocity and heading angle plots, we can see a discontinuity starting from $t = 1.2s$. After that, the simulated agent starts to deviate from the road. In another example shown in Fig. 3(b), the simulated agent slows down and stops in the middle of the roundabout. Similarly, we can observe discontinuity in velocity and heading angle profile. 
\begin{figure}[htbp]
\centering
\minipage{0.48\textwidth}
  \includegraphics[width=1\linewidth]{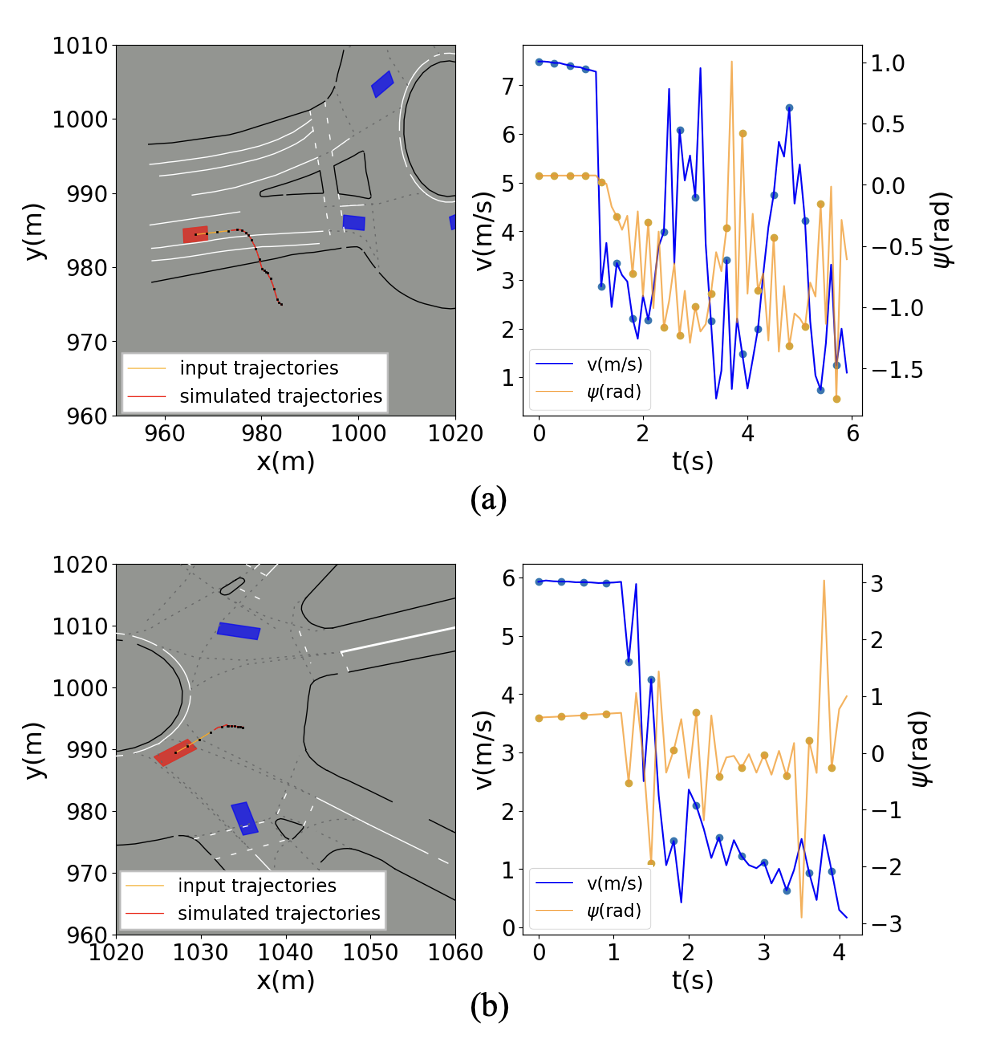}
  \caption{Visualization of failure cases in \emph{xy} setting: (a) Entering roundabout scenarios; (b) leaving roundabout scenarios. The simulated agent loses its stability after a few steps. The velocity and angle profile show the discontinuity of states in closed-loop simulation. For visualization purposes, the dots are plotted every 0.3 seconds.}
\endminipage\hfill
\end{figure}

\begin{figure*}
    \centering
     \includegraphics[width=\linewidth]{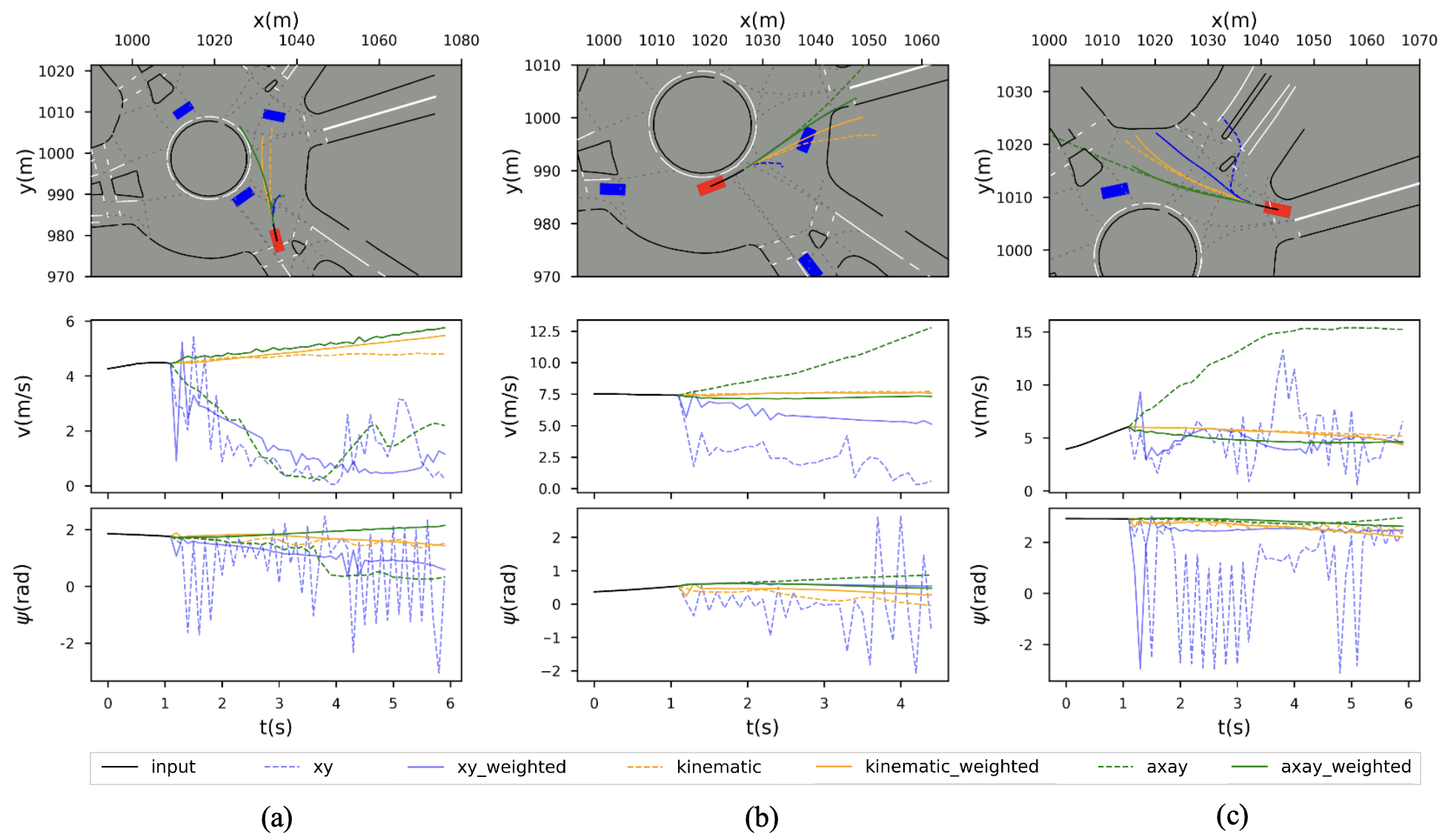}
    \caption{Ablation study of kinematic layer and smoothing layer. We input a 1-second historical states to simulate the agent for 5 seconds. The simulated trajectories, velocities, and angle profiles are shown in the figure. The figure shows that \emph{axay weighted} performs the best in all scenarios. We can observe that the heading angle $\psi$ profile of \emph{kinematic} has oscillation in (a) and (c), which implies that the kinematic bicycle layer is sensitive to the initial heading angle.
    }
    \label{kinematic bicycle model}
\end{figure*} 
Based on our observation, we find that the stability of the closed-loop simulation is strongly related to the consistency between the predicted and past sequences. Once the ego vehicle takes a step that does not smoothly connect with the past state sequences, it becomes difficult for the model to recover from the resulted state deviation, which cause the model generates either off-road or jerky trajectory. Therefore, how to maintain the consistency between the simulated states and historical states becomes a crucial problem when running closed-loop simulations.

\subsection{Ablation Study}
In this subsection, we compare six different framework settings: \emph{xy}, \emph{xy weighted}, \emph{kinematic}, \emph{kinematic weighted}, \emph{axay}, and \emph{axay weighted}. The influences of adding the kinematic layer and the weighted average smoothing layer to the model are discussed. Furthermore, differences between kinematic bicycle layer \emph{kinematic} and simplified kinematic layer \emph{axay} are also discussed.

One key observation we find is that when the velocity $v$ and heading angle $\psi$ do not vary much during the process, the closed-loop simulation will gradually deviate from the road instead of performing unreasonable behaviors right after a few steps. In Fig. 4(a), \emph{xy} simulates the ego vehicle stopping in the middle of the roundabout with jerky velocity and angle profile. On the other hand, \emph{kinematic}, \emph{kinematic weighted}, and \emph{axay weighted} all simulate reasonable trajectories with smooth state sequences. Similar observations can be seen in both Fig. 4(b) and Fig. 4(c).

\begin{table*}
\centering
\caption{\label{tab:table-name} Comparison between six different framework settings. The prediction metrics reflect the prediction performance of each predictor (open-loop). The closed-loop evaluation metrics reflect simulation performance, where each simulation is simulated for 5 seconds. Note that for prediction metrics, it is unnecessary to add a weighted average layer because the predictors are not evaluated in closed-loop.}
\resizebox{\textwidth}{!}{%
\renewcommand{\arraystretch}{1.5}
\begin{tabular}{l|ccc|cccccccccc}
\hline
\multicolumn{1}{c|}{\multirow{3}{*}{Model}} & \multicolumn{3}{c|}{Prediction Metrics}                                                                                                                        & \multicolumn{10}{c}{Closed-Loop Simulation Metrics}                                                                                                                                                                                                                                                                                                                                         \\ \cline{2-14} 
\multicolumn{1}{c|}{}                       & \multicolumn{1}{c|}{\multirow{2}{*}{\thead{ADE \\(m)}}}              & \multicolumn{1}{c|}{\multirow{2}{*}{\thead{FDE \\(m)}}}             & \multirow{2}{*}{\thead{MR \\($\%$)}}              & \multicolumn{5}{c|}{ADE (m)}                                                                                                              & \multicolumn{1}{c|}{\multirow{2}{*}{\thead{Overall \\ADE (m)}}} & \multicolumn{1}{c|}{\multirow{2}{*}{\thead{FDE \\(m)}}} & \multicolumn{1}{c|}{\multirow{2}{*}{\thead{CR \\($\%$)}}} & \multicolumn{1}{c|}{\multirow{2}{*}{\thead{MS \\(m/$s^{3}$)}}} & \multirow{2}{*}{\thead{TD \\(m)}} \\ \cline{5-9}
\multicolumn{1}{c|}{}                       & \multicolumn{1}{c|}{}                                      & \multicolumn{1}{c|}{}                                     &                                       & 0-1s                   & 1-2s                  & 2-3s                   &3-4s                   & \multicolumn{1}{c|}{4-5s}                    & \multicolumn{1}{c|}{}                                 & \multicolumn{1}{c|}{}                         & \multicolumn{1}{c|}{}                         & \multicolumn{1}{c|}{}                                             &                         \\ \hline
\emph{xy}                                          & \multicolumn{1}{c|}{\multirow{2}{*}{0.57 (0.19)}}          & \multicolumn{1}{c|}{\multirow{2}{*}{\textbf{1.41 (0.4)}}} & \multirow{2}{*}{20.6 (4.6)}          & 4.37 (0.59)          & 10.32 (1.39)        & 15.78 (2.19)         & 21.74 (2.68)         & \multicolumn{1}{c|}{27.68 (2.46)}          & \multicolumn{1}{c|}{15.56 (1.81)}                     & \multicolumn{1}{c|}{22.78 (1.87)}             & \multicolumn{1}{c|}{32.0 (11.0)}              & \multicolumn{1}{c|}{207.65 (26.22)}                               & 37.51 (4.02)            \\
\emph{xy weighted}                                 & \multicolumn{1}{c|}{}                                      & \multicolumn{1}{c|}{}                                     &                                       & 3.41 (0.79)          & 6.63 (1.58)         & 10 (2.01)            & 13.8 (2.36)          & \multicolumn{1}{c|}{17.89 (2.65)}          & \multicolumn{1}{c|}{10.08 (1.83)}                     & \multicolumn{1}{c|}{14.82 (2.07)}             & \multicolumn{1}{c|}{24.0 (6.1)}              & \multicolumn{1}{c|}{73.42 (22.75)}                                & 27.37 (3.69)            \\ \hline
\emph{kinematic}                                   & \multicolumn{1}{c|}{\multirow{2}{*}{\textbf{0.55 (0.11)}}} & \multicolumn{1}{c|}{\multirow{2}{*}{1.46 (0.29)}}         & \multirow{2}{*}{\textbf{19.3 (2.6)}} & 2.28 (0.25)          & 5.43 (1.65)         & 9.51 (3.26)          & 14.03 (4.38)         & \multicolumn{1}{c|}{18.68 (4.85)}          & \multicolumn{1}{c|}{9.69 (2.74)}                      & \multicolumn{1}{c|}{15.23 (4.09)}             & \multicolumn{1}{c|}{22.0 (6.7)}              & \multicolumn{1}{c|}{\textbf{2.51 (0.76)}}                         & \textbf{11.52 (3.89)}            \\
\emph{kinematic weighted}                          & \multicolumn{1}{c|}{}                                      & \multicolumn{1}{c|}{}                                     &                                       & 2.27 (0.2)           & 4.59 (0.52)         & 7.49 (0.93)          & 11.11 (1.46)         & \multicolumn{1}{c|}{15.07 (2.09)}          & \multicolumn{1}{c|}{7.88 (0.98)}                      & \multicolumn{1}{c|}{12.87 (1.74)}             & \multicolumn{1}{c|}{18.0 (2.4)}              & \multicolumn{1}{c|}{8.61 (3.86)}                                  & 11.95 (2.7)             \\ \hline
\emph{axay}                                  & \multicolumn{1}{c|}{\multirow{2}{*}{0.65 (0.03)}}          & \multicolumn{1}{c|}{\multirow{2}{*}{1.79 (0.09)}}         & \multirow{2}{*}{21.3 (2.9)}          & 1.87 (0.06)          & 5.08 (0.13)         & 9.67 (0.26)          & 15.44 (0.51)         & \multicolumn{1}{c|}{21.92 (0.84)}          & \multicolumn{1}{c|}{10.76 (0.35)}                     & \multicolumn{1}{c|}{16.21 (0.61)}             & \multicolumn{1}{c|}{27.0 (7.3)}              & \multicolumn{1}{c|}{8.49 (3.87)}                                  & 74.04 (19.55)           \\
\emph{axay weighted}                          & \multicolumn{1}{c|}{}                                      & \multicolumn{1}{c|}{}                                     &                                       & \textbf{1.67 (0.05)} & \textbf{3.2 (0.12)} & \textbf{5.37 (0.05)} & \textbf{8.17 (0.14)} & \multicolumn{1}{c|}{\textbf{11.35 (0.36)}} & \multicolumn{1}{c|}{\textbf{5.93 (0.06)}}             & \multicolumn{1}{c|}{\textbf{9.87 (0.17)}}     & \multicolumn{1}{c|}{\textbf{14.0 (0.9)}}     & \multicolumn{1}{c|}{11.1 (3.22)}                                  & {47.91 (3.47)}   \\ \hline
\end{tabular}
}
\end{table*}


\subsubsection{Sensitivity of the Kinematic Layer}
Since the kinematic layer calculates future trajectories based on the velocity $v$ and heading angle $\psi$, it is susceptible to the input initial states. For models that incorporate the kinematic bicycle layer (i.e., \emph{kinematic} and \emph{kinematic weighted}), both the heading angle and velocity are crucial to the predicted sequences. For example, in Fig. 4(a) and 4(c), we can see there is an obvious oscillating of the heading angle of the simulated agent in \emph{kinematic}. We believe that the sensitivity issue of the kinematic layer may cause the simulated trajectories to slightly deviate from the road, as shown in Fig. 4(b) and 4(c). Similarly, \emph{axay} is sensitive to the initial velocities $v_x$ and $v_y$. For example, in Fig. 4(b) and 4(c), when the model \emph{axay} predicts a faster speed, the simulated agent keeps accelerating and eventually deviate from the road.

\subsubsection{Effect of the Weighted Average Method}
The weighted average method prevents the simulated agent from making a sudden change of trajectories and thus improves the stability of the model. The effect can be seen in the improvement from \emph{xy} to \emph{xy weighted} setting. In Fig. 4(b), \emph{xy weighted} successfully simulates the agent entering the branch. In Fig. 4(c), the simulated agent in \emph{xy} suddenly drives toward the divisional island with a jerky trajectory, while \emph{xy weighted} gradually deviates from the road.
In both cases, the velocity and heading angle profile in \emph{xy weighted} do not have a dramatic oscillation compared to \emph{xy}. As for framework settings with the kinematic bicycle layer, the simulated agent under the \emph{kinematic} setting performs similarly to that under the \emph{kinematic weighted} setting.

The weighted average method also maintains stability in simplified kinematic layer setting \emph{axay}. From Fig. 4(a), the simulated agent in \emph{axay} will either accelerate to drive off-road (Fig. 4(b) and Fig. 4(c)) or decelerate and deviate from typical trajectories (Fig. 4(a)). We argue that the weighted average helps the predicting state sequences maintain in a certain range that the model can preserve its stability. This characteristic allows the 
closed-loop simulation framework to have a tolerance for slightly deviated predictions. However, this is not a perfect solution for closed-loop simulation since the weighted average will introduce inertia to the system. Thus, the simulated agent cannot react immediately to other vehicles.

Among all six different framework settings, \emph{axay weighted} performs the best in terms of stability and reasonableness. We believe there are three main reasons. First, using the kinematic layer will produce a smoother state sequence to reduce the discrepancy between the historical and simulated states. Second, in terms of sensitivity, \emph{axay} is only sensitive to velocities, while 
\emph{kinematic} is sensitive to both velocities and heading angles. We argue that the heading angle will have more influence on the simulated trajectories, and this may be the reason why \emph{axay weighted} works better than \emph{kinematic}. At last, the weighted average method allows the closed-loop simulation framework to have a tolerance for predictions that deviate from normal trajectories.

\subsection{Quantitative Analysis of Proposed Approach}
Table 1 shows the comparison of six different framework settings in randomly selected 100 scenarios. Although the prediction performance under setting \emph{axay} is slightly worse than that under setting \emph{xy} and \emph{kinematic}, \emph{axay weighted} outperforms other framework settings overall in closed-loop simulation metrics. Also, according to the standard deviation of all these metrics, \emph{axay weighted} has the most consistent performances. In terms of motion smoothness, \emph{xy} has the largest MS metric with 207.65 m/$s^{3}$. This large MS value reflects that directly using the predictor to simulate trajectories will easily cause instability and produce jerky trajectories. After adding a kinematic layer, we can observe that MS is greatly improved in both \emph{axay} and \emph{kinematic} settings. 

Compared with \emph{xy}, the overall performance of \emph{xy weighted} is greatly improved, especially for the motion smoothness metric. Similarly, we can also observe the improvement of \emph{axay} compared to \emph{axay weighted} under most metrics. Note that the TD value for both \emph{xy} and \emph{axay} settings is dropped in terms of mean and standard deviation after adding the weighted average method. This implies that the weighted average improves the consistency of the predictions at each timestamp and thus improves the closed-loop simulation performance. On the other hand, for \emph{kinematic} to \emph{kinematic weighted} settings, both MS and TD do not have improvements, while simulation realism (ADE and FDE) and collision rate (CR) are still improved. One potential reason is that the weighted average smoothing layer improves the model's tolerance when the model produces a deviated state sequence. In addition, the ADE and FDE of \emph{kinematic} and \emph{kinematic weighted} are larger than those of \emph{axay weighted}. We believe this result comes from the sensitivity issue of the kinematic bicycle model: a slight angle deviation would cause the simulated agent to accumulate its error during closed-loop simulation.
\subsection{Further Discussion}
Currently, there are still a few cases that will gradually drive off-road even in our best framework setting, \emph{axay weighted} (shown in Fig. 5). Most of these failure cases are under a situation that will easily cause model confusion. For example, in Fig. 5(a), the simulated agent has two driving options: one is to leave the roundabout and drive towards an exit branch, and the other is to stay in the roundabout. If we use a deterministic neural network model such as Vectornet to predict the agent's trajectory, the model will tend to pick the average of two plans in order to minimize the training loss. Therefore, introducing multi-modality to the agents' future plans will resolve this issue.  Another potential reason is that the state error of the simulated vehicle starts to accumulate when we simulate the agent in closed-loop. At each time step, the simulated agent takes actions based on slightly deviated trajectories and eventually drives off-road. This is a common distributional shift problem in imitation learning. In the future, how to deal with distribution shift would be an important topic in closed-loop simulation for long-horizon simulation. In general, it is common to assume a high-performance predictor would also perform well under the closed-loop setting. However, in this work, we discover that it is not the case, and both the smoothness and consistency of the simulated state sequences are important to the stability of closed-loop simulation. 



\begin{figure}[htbp]
\centering
\minipage{0.5\textwidth}
  \includegraphics[width=\linewidth]{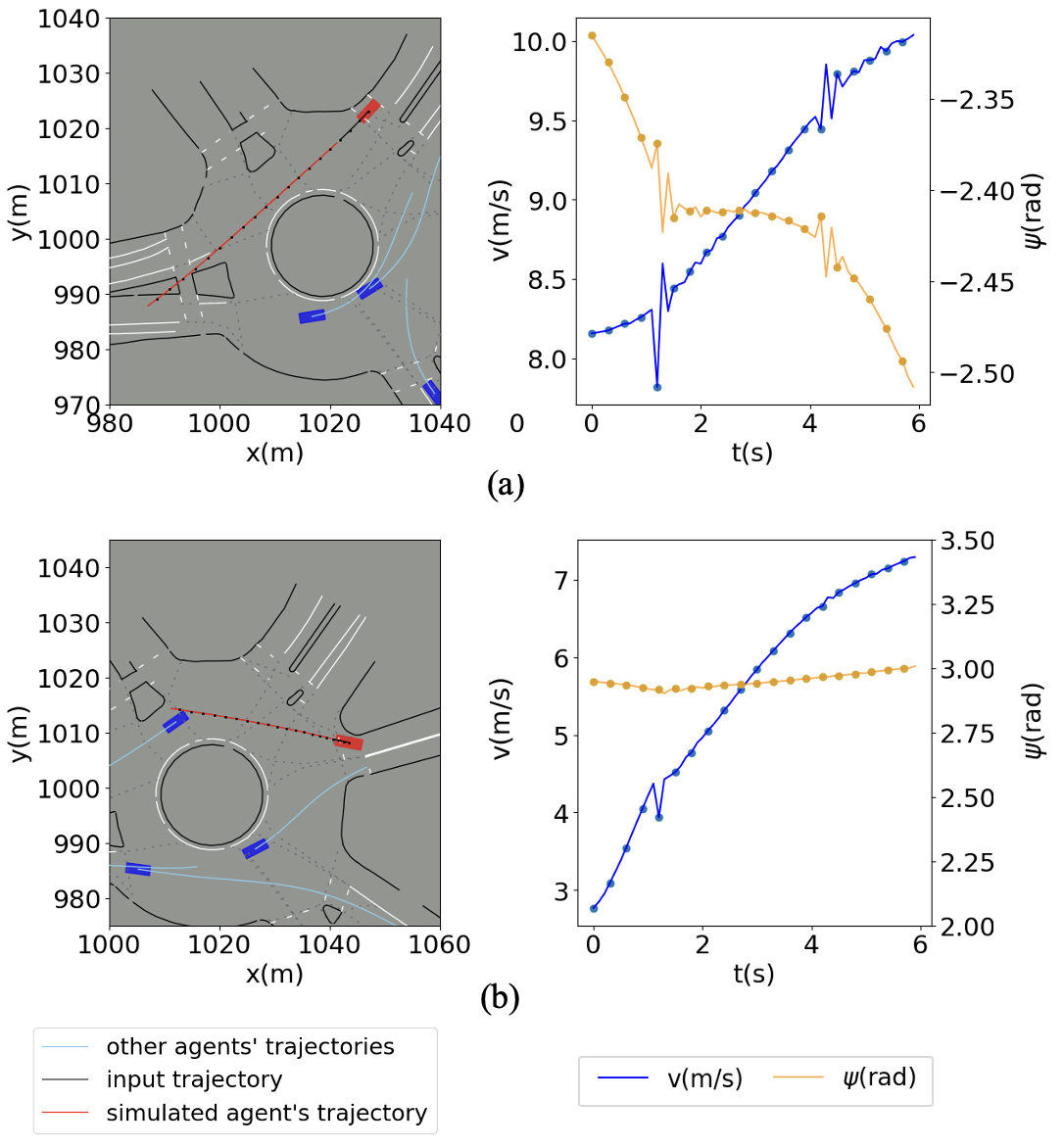}
\endminipage\hfill
\caption{Visualization of simulated agent's trajectories, velocity, and angle profile. This figure shows the distribution shift problem in the long horizon. For visualization purposes, the dots are plotted every 0.3 seconds.}
\end{figure}

\section{CONCLUSION}

In this paper, a novel simulation framework and a thorough stability analysis of reactive simulation are presented. Our experiments show that directly using a predictor in the reactive simulation will easily generate off-road or jerky trajectories. We find out that during the closed-loop simulation, the smoothness and consistency of the simulated state sequences are crucial to stability. The kinematic bicycle layer can improve the stability but suffer from the sensitivity issue of initial states. On the other hand, the simplified kinematic layer performs the best in terms of simulation realism and stability. Furthermore, we demonstrate how our weighted average method improves the stability in reactive simulation. For future work, we will tackle the long horizon distribution shift problem and consider the multi-modality of the traffic scenarios.

\bibliography{iros2021.bib}
\bibliographystyle{IEEEtran}
\end{document}